\pdfoutput=1

\documentclass[11pt]{article}

\usepackage[]{acl}

\usepackage{times}
\usepackage{latexsym}

\usepackage[T1]{fontenc}

\usepackage[utf8]{inputenc}

\usepackage{microtype}
\usepackage{graphicx}
\usepackage{multirow}
\usepackage{makecell}
\usepackage{amsmath}
\usepackage{amsfonts}
\usepackage{subfigure}
\usepackage{booktabs}
\usepackage{hyperref}

\title{A Sentence is Worth 128 Pseudo Tokens: A Semantic-Aware Contrastive Learning Framework for Sentence Embeddings}

\author{Haochen Tan$^{1, 2}$, Wei Shao$^{1, 2}$, Han Wu$^{1, 2}$, Ke Yang$^{3}$, Linqi Song$^{1,2}\thanks{~~Corresponding author.}$\\
$^{1}$City University of Hong Kong Shenzhen Research Institute \\
$^{2}$Department of Computer Science, City University of Hong Kong\\
$^{3}$School of Integrated Circuits, Peking University \\
\texttt{haochetan2-c@my.cityu.edu.hk}\\
\texttt{linqi.song@cityu.edu.hk}\\
  }

\begin{document}
\maketitle
\begin{abstract}
Contrastive learning has shown great potential in unsupervised sentence embedding tasks, e.g., SimCSE \citep{gao2021simcse}.
However, We find that these existing solutions are heavily affected by superficial features like the length of sentences or syntactic structures. In this paper, we propose a semantics-aware contrastive learning framework for sentence embeddings, termed Pseudo-Token BERT (PT-BERT), which is able to exploit the pseudo-token space (i.e., latent semantic space) representation of a sentence while eliminating the impact of superficial features such as sentence length and syntax. Specifically, we introduce an additional pseudo token embedding layer independent of the BERT encoder to map each sentence into a sequence of pseudo tokens in a fixed length. Leveraging these pseudo sequences, we are able to construct same-length positive and negative pairs based on the attention mechanism to perform contrastive learning. In addition, we utilize both the gradient-updating and momentum-updating encoders to encode instances while dynamically maintaining an additional queue to store the representation of sentence embeddings, enhancing the encoder's learning performance for negative examples. Experiments show that our model outperforms the state-of-the-art baselines on six standard semantic textual similarity (STS) tasks. Furthermore, experiments on alignments and uniformity losses, as well as hard examples with different sentence lengths and syntax, consistently verify the effectiveness of our method.

\end{abstract}

\section{Introduction}
Sentence embedding serves as an essential technique in a wide range of applications, including semantic search, text clustering, text classification, etc.~\cite{skipthought, quickthought, infersent, universalsent, sbert,gao2021simcse}. Contrastive learning works on learning representations such that similar examples stay close whereas dissimilar ones are far apart, and thus is suitable for sentence embeddings due to its natural availability of similar examples. Incorporating contrastive learning in sentence embeddings improves the efficiency of semantic information learning in an unsupervised manner \cite{MoCo, simCLR} and has been shown to be effective on a variety of tasks~\cite{sbert, gao2021simcse, zhang-etal-2020-unsupervised}. 

\begin{figure}[t!]
    \centering
    \begin{minipage}[t]{0.5\textwidth}
    \includegraphics[width=0.97\linewidth]{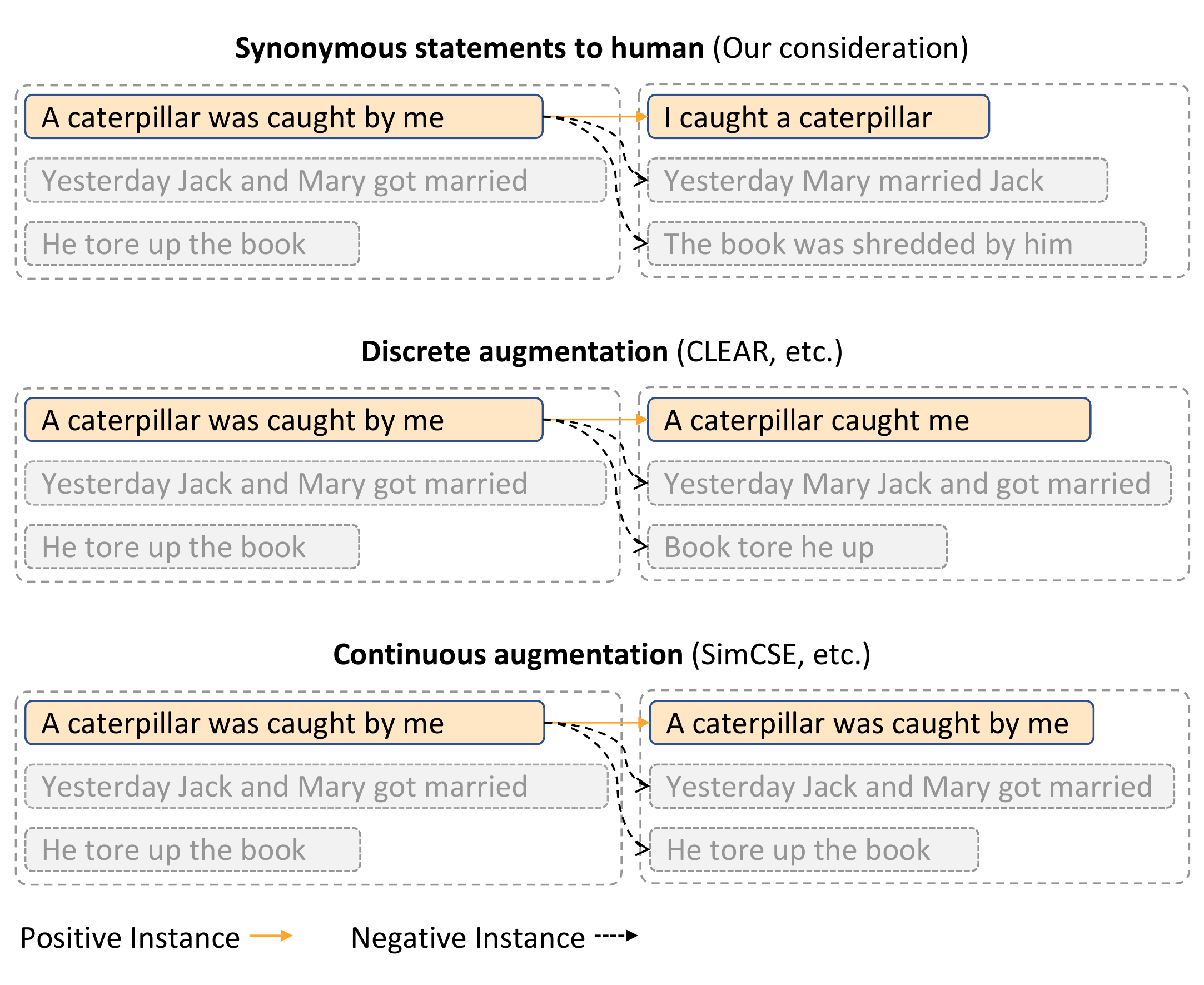}
    \end{minipage}
    \caption{A realistic scenario is described at the top, negative examples have the same length and structure, while positive examples act in the opposite way. In comparison, discrete augmentation obtains positive instances with word deletion or reordering~\cite{wu2020clear,meng2021cocolm}, which may misinterpret the meaning. The continuous method treats embeddings of the same original sentence as positive examples and augments sentences with the different encoding functions~\cite{Carlsson2021SemanticRW, gao2021simcse}. }
    \label{fig:motivation}
\end{figure}


In contrastive learning for sentence embeddings, a key challenge is constructing positive instances. Both discrete and continuous augmentation methods have been studied recently. Methods in~\citet{wu2018unsupervised,meng2021cocolm} perform discrete operations directly on the original sentences, such as word deletion and sentence shuffling, to get positive samples. However, these methods may lead to unacceptable semantic distortions or even complete misinterpretations of the original statement. In contrast, the SimCSE method~\cite{gao2021simcse} obtains two different embeddings in the continuous embedding space as a positive pair for one sentence through different \textit{dropout masks}~\cite{dropout} in the neural network for representation learning. Nonetheless, this method overly relies on superficial features existing in the dataset like sentence lengths and syntactic structures and may pay less reflection on meaningful semantic information. As an illustrative example, the sentence-pair in Fig.~\ref{fig:motivation} ``A caterpillar was caught by me.'' and ``I caught a caterpillar.'' appear to organize differently in expression but convey exactly the same semantics. 

To overcome these drawbacks, in this paper, we propose a semantic-aware contrastive learning framework for sentence embeddings, termed Pseudo-Token BERT (PT-BERT), that is able to capture the pseudo-token space (i.e., latent semantic space) representation while ignoring effects of superficial features like sentence lengths and syntactic structures. 
Inspired by previous works on prompt learning and sentence selection~\cite{li2021prefixtuning,liu2021gpt,polyencoder}, which create a pseudo-sequence and have it serve the downstream tasks, we present PT-BERT to train pseudo token representations and then to map sentences into pseudo token spaces based on an attention mechanism. 

In particular, we train additional 128 pseudo token embeddings, together with sentence embeddings extracted from the BERT model (i.e., gradient-encoder), and then use the attention mechanism~\cite{NIPS2017_3f5ee243} to map the sentence embedding to the pseudo token space (i.e., semantic space). We use another BERT model (i.e., momentum-encoder~\cite{MoCo}) to encode the original sentence, adopt a similar attention mechanism with the pseudo token embeddings, and finally output a continuously augmented version of the sentence embedding. We treat the representations of the original sentence encoded by the gradient-encoder and the momentum-encoder as a positive pair. In addition, the momentum-encoder also generates negative examples, dynamically maintains a queue to store these negative examples, and updates them over time. By projecting all sentences onto the same pseudo sentence, the model greatly reduces the dependence on sentence length and syntax when making judgments and makes the model more focused on the semantic level information. 

In our experiments, we compare our results with the previous state-of-the-art work. We train PT-BERT on $10^6$ randomly sampled sentences from English Wikipedia and evaluate on seven standard semantic textual similarity (STS) tasks~\cite{agirre-etal-2012-semeval,agirre-etal-2013-sem,agirre-etal-2014-semeval,agirre-etal-2015-semeval,agirre-etal-2016-semeval}~\cite{DBLP:conf/lrec/MarelliMBBBZ14}. Besides, we also compare our approach with a framework based on an advanced discrete augmentation we proposed. We obtain a new state-of-the-art on standard semantic textual similarity tasks with our PT-BERT, which achieves $77.74\%$ of Spearman's correlation. To show the effectiveness of pseudo tokens, we calculate the align-loss and uniformity loss~\cite{wang2020understanding} and verify our approach on a sub-dataset with hard examples sampled from STS-(2012-2016). We have released our source code at \url{https://github.com/Namco0816/PT-BERT} to facilitate future work. 
\section{Related Work}
In this section, we discuss related studies with repect to the contrastive learning framework and sentence embedding.
\subsection{Contrastive Learning for Sentence Embedding}
\paragraph{Contrastive learning.}
Contrastive learning ~\cite{1640964} has been used with much success in both natural language processing and computer vision~\cite{yang-etal-2019-reducing, klein-nabi-2020-contrastive, simCLR, MoCo, gao2021simcse}. In contrast to generative learning, contrastive learning requires learning to distinguish and match data at the abstract semantic level of the feature space. It focuses on learning common features between similar examples and distinguishing differences between non-similar examples. In order to compare the instances with more negative examples and less computation, memory bank~\cite{wu2018unsupervised} is proposed to enhance the performance under the contrastive learning framework. While with a large capacity to store more samples, the memory bank is not consistent enough, which could not update the negative examples during comparison. Momentum-Contrast (MoCo)~\cite{MoCo} uses a queue to maintain the dictionary of samples which allows the model to compare the query with more keys for each step and ensure the consistency of the framework. It updates the parameter of the dictionary in a momentum way. 

\paragraph{Discrete and continuous augmentation.}
By equipping discrete augmentation that modifies sentences directly on token level with contrastive learning, significant success has been achieved in obtaining sentence embeddings. Such methods include word omission~\cite{yang-etal-2019-reducing}, entity replacement~\cite{DBLP:conf/iclr/XiongDWS20}, trigger words~\cite{klein-nabi-2020-contrastive} and traditional augmentations such as deletion, reorder and substitution~\cite{wu2020clear,meng2021cocolm}. Examples with diverse expressions can be learned during training, making the model more robust to expressions of different sentence lengths and styles. However, these approaches are limited because there are huge difficulties in augmenting sentences precisely since a few changes can make the meaning completely different or even opposite. 

Researchers have also explored the possibility of building sentences continuously, which instead applies operation in embedding space. CT-BERT~\cite{Carlsson2021SemanticRW} encodes the same sentence with two different encoders. Unsup-SimCSE~\cite{gao2021simcse} compares the representations of the same sentence with different dropout masks among the mini-batch. These approaches continuously augment sentences while retaining the original meaning. However, positive pairs seen by SimCSE always have the same length and structure, whereas negative samples are likely to act oppositely. As a result, sentence length and structure are highly correlated to the similarity score of examples. During training, the model has never seen positive samples with diverse expressions, so that in real test scenarios, the model would be more inclined to classify the synonymous pairs with different expressions as negatives, and those sentences with the same length and structures are more likely to be grouped as positive pairs. This may cause a biased encoder.

\subsection{Pseudo Tokens} 
In the domain of prompt learning~\cite{liu2021gpt,10.1162/tacl_a_00324, li2021prefixtuning,DBLP:journals/corr/abs-2012-15723}, the way to create prompt can be divided into two types, namely discrete and continuous ways. Discrete methods usually search the natural language template as the prompt~\cite{davison-etal-2019-commonsense,petroni-etal-2019-language}, while the continuous way always directly works on the embedding space with "pseudo tokens"~\cite{liu2021gpt, li2021prefixtuning}. In retrieval and dialogue tasks, the current approach adopts "pseudo tokens", namely "poly codes" ~\cite{polyencoder}, to jointly encode the query and response precisely and ensure the inference time when compared with the Cross-Encoders and Bi-Encoders~\cite{wolf2019transfertransfo, mazare-etal-2018-training,dinan2019wizard}. The essence of these methods is to create a pseudo-sequence and have it serve the downstream tasks without the need for humans to understand the exact meaning. 
The parameters of these pseudo tokens are independent of the natural language embeddings, and can be tuned based on a specific downstream task. In the following sections, we will show the idea to weaken the model's consideration of sentence length and structures by introducing additional pseudo token embeddings on top of the BERT encoder. 

\begin{figure*}[ht!]
    \centering
    \includegraphics[width=1.0\linewidth]{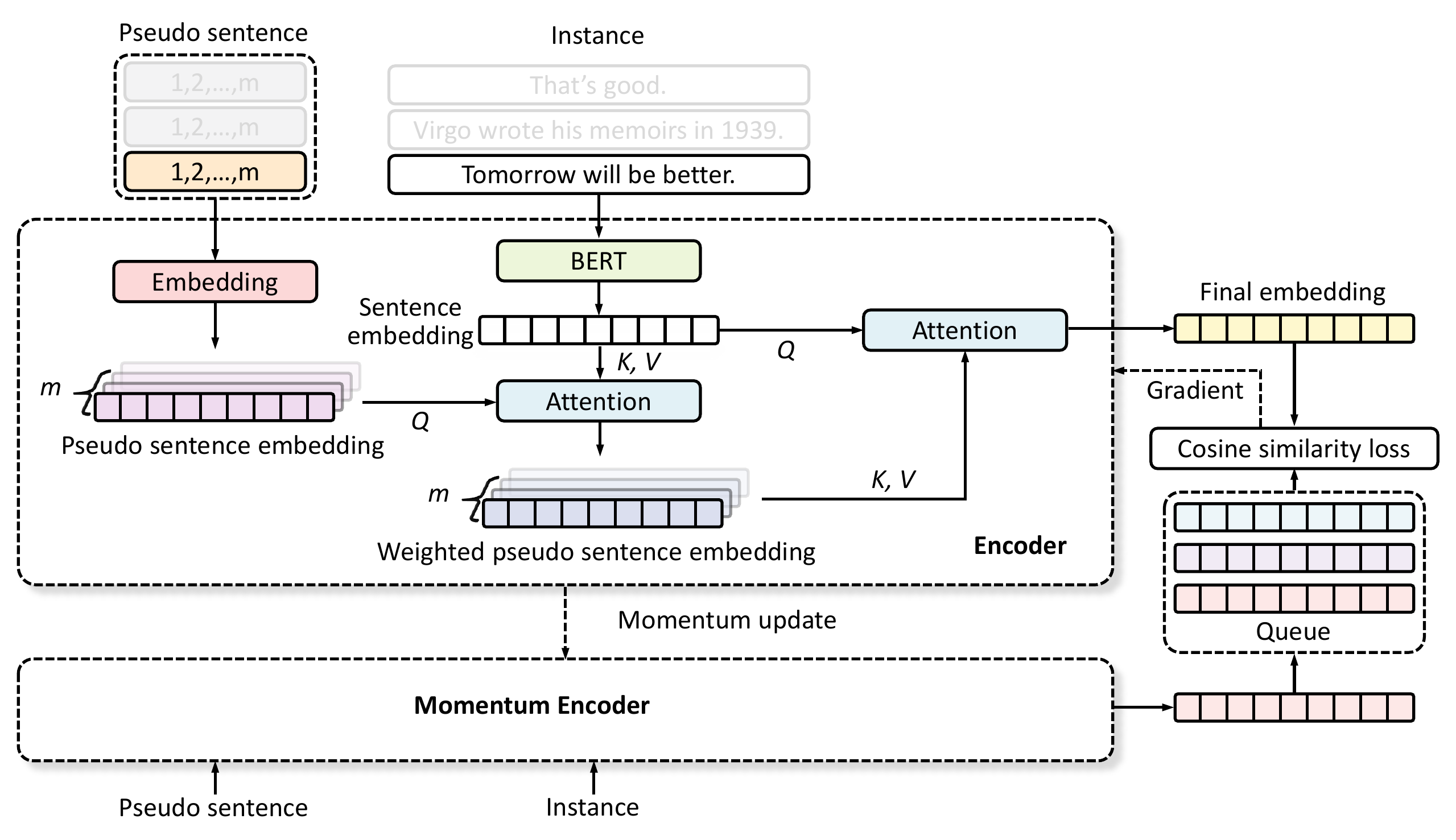}
    \caption{The model is divided into two parts, the upper part (Encoder) updates the learnable parameters with gradient, while the bottom (Momentum Encoder) inherits parameters from the upper part with momentum-updating. We repeatedly input the same sequence of pseudo tokens while processing the original sentences. An additional BERT attention mapping the pooler-output of BERT to pseudo sequence representation, extending the sentence embedding to a fixed length and mapping the syntactic structure to the style of the pseudo sentence. The two attentions in the figure are the same and with identical parameters.}
    \label{fig:architecture}
\end{figure*}

\section{Methods}
In this section, we introduce PT-BERT, which provides novel contributions on combining advantages of both discrete and continuous augmentations to advance the state-of-art of sentence embeddings. We first present the setup of problems with a thorough analysis on the bias introduced by the textual similarity theoretically and experimentally. Then we show the details of Pseudo-Token representation and our model's architecture.
\subsection{Preliminary}
\label{sec:longshortpairs}

\begin{table}
\centering
\begin{tabular}{lcc}
\hline
 &\textbf{Sub-dataset} & \textbf{original} \\
\hline
STS12 &66.54 &68.40\\
STS13 &78.50 &82.41\\
STS14 &68.76 &74.38\\ 

STS15 &70.27 &80.91\\ 
STS16 &71.31 &78.56\\

\hline
\end{tabular}
\caption{SimCSE's results on sub-dataset from STS12-16, comparing with original results.}
\label{tab:longshortpairs}
\end{table}

\begin{table}
\centering
\begin{tabular}{lccc}
\hline
 &{SimCSE$_{32}$} & {SimCSE$_{64}$} &SimCSE$_{128}$ \\
\hline
Avg.& 76.25 & 75.20 & 75.29 \\
\hline
\end{tabular}
\caption{Different acceptable sequence length of SimCSE would affect the result on STS tasks.}
\label{tab:longshorttrain}
\end{table}

Consider a sentence $s$, we say that the augmentation is continuous if $s$ is augmented by different encoding functions, $f(\cdot)$ and $f'(\cdot)$. Sentence embeddings $\textbf{h} = f(s)$ and $\textbf{h}' = f'(s)$ are obtained by these two functions. With a slight change of the encoding function (e.g., encoders with different \textit{dropout} masks), $\bf h'$ can be seen as a more precisely augmented version of $\bf h$ compared with the discrete augmentation. Semantic information of $\bf h'$ should be the same as $\bf h$. Therefore, $\bf h$ and $\bf h'$ are a pair of positive examples and we could randomly sample a sentence to construct negative example pairs. 

Previous state-of-the-art models ~\cite{gao2021simcse} adopt the continuous strategy that augments sentences with \textit{dropout}~\cite{dropout}. It is obvious that all the positive examples in SimCSE have the same length and structure while negative examples act oppositely. In this way, SimCSE will inevitably take these two factors as hints during test. To further verify this conjecture, we sort out the positive pairs with a length difference of more than five words and negative pairs of less than two words from STS-(2012-2016). 

Table~\ref{tab:longshortpairs} shows that the performance of SimCSE plummets on this dataset. Besides, we also find that SimCSE truncates all training corpus into 32 tokens, which shortens the discrepancy of the sentence's length. After we scale the max length that SimCSE could accept from 32 to 64 and 128, the performance degrades significantly during the test even though the model is supposed to learn more from the complete version of sentences(See Table~\ref{tab:longshorttrain}). The reason for this result may lie in the fact that, without truncation, all positive  pairs still have the same length, whereas the difference in length between the negative and positive ones is 
enlarged. Therefore, the encoder will rely more on sentence length and make the wrong decision.

\subsection{Pseudo-Token BERT}
We realize it is vital to train an unbiased encoder that captures the semantics and also would not introduce intermediate errors. This motivates us to propose the PT-BERT, as evidence shows that the encoder may fail to make predictions when trained on a biased dataset with same-length positive pairs, by learning the spurious correlations that work only well on the training dataset~\cite{Arjovsky2019InvariantRM,nam2020learning}. 
\paragraph{Pseudo-Token representations.}
The idea of PT-BERT is to reduce the model's excessive dependence on textual similarity when making predictions. Discrete augmentation achieves this goal by providing both positive and negative examples with diverse expressions. Therefore the model does not jump to conclusions based on sentence length and syntactic structure during the test. 

Note that we achieve this same purpose in a seemingly opposite way: \textit{mapping the representations of both positive and negative examples to a pseudo sentence with the same length and structure.}  We take an additional embedding layer outside the BERT encoder to represent a pseudo sentence \{$0, 1,...,m$\} with fixed length $m$ and syntax. This embedding layer is fully independent of the BERT encoder, including the parameters and corresponding vocabulary. Random initialization is applied to this layer, and each parameter will be updated during training. The size of this layer depends on the vocabulary of pseudo tokens(length of pseudo sentences). Besides, adopting the attention mechanism \citep{NIPS2017_3f5ee243, Bahdanau2015NeuralMT,pmlr-v70-gehring17a}, we take the pseudo sentence embeddings as the query states of cross attention while key and value states are the sentence embeddings obtained from the BERT encoder. This allows the pseudo sentence to attend to the core part and ignore the redundant part of original sentence while keeping the fixed length and structure.

Fig.~\ref{fig:architecture} illustrates the framework of PT-BERT. Denoting the pseudo sentence embedding as $\textbf{P}$ and the sentence embedding encoded by BERT as $\textbf{Y}$, we obtain the weighted pseudo sentence embedding of each sentence by mapping the sentence embedding to the pseudo tokens with attention:

\begin{gather}
    \bf Z_i' = \rm{Attention}(\bf PW^Q, \bf Y_iW^K, \bf Y_iW^V) \\
    \rm{Attention}(\bf Q, \bf K, \bf V) = \rm{softmax}(\frac{\bf QK^T}{\sqrt{d_k}})\bf V, 
\end{gather}
where $d_k$ is the dimension of the model, $\bf W^Q$, $\bf W^K$, $\bf W^V$ are the learnable parameters with $\mathbb{R}^{d_k \times d_k}$, $i$ denotes the $i$-th sentence in the dataset. Then we obtain the final embedding ${\bf h_i}$ with the same attention layer by mapping pseudo sentences back to original sentence embeddings:
\begin{gather}
    \bf h_i = \rm{Attention}(\bf Y_iW^Q,\bf Z_i'W^K,\bf Z_i'W^V). 
\end{gather}

Finally, we compare the cosine similarities between the obtained embeddings of $\bf h$ and $\bf h'$ using Eq.~\ref{for:loss_our} , where $\bf h'$ are the samples encoded by the momentum-encoder and stored in a queue.

\begin{table*}[!ht]
    \centering
    \begin{tabular}{lcccccccc}
    \toprule[0.8pt]
    \bf Model &  \bf STS12 & \bf STS13 & \bf STS14 & \bf STS15 & \bf STS16 & \bf STS-B & \bf SICK-R & \bf Avg. \\
    \hline
    \multicolumn{9}{c}{\textit{Discrete Augmentation}} \\
    \hline
    CLEAR&49.00&48.90&57.40&63.60&65.60&72.50&\bf 75.60&61.80 \\
    MoCo&68.35&81.42&73.34&{81.63}&78.61&76.40&68.50&75.46\\
    MoCo+reorder& 66.14 & 80.06 & 73.14 & 81.35 & 76.01 & 73.99 & 65.76 & 73.78\\
    MoCo+duplication&65.88&{82.24}&73.34&81.49&77.48&76.29&68.86&75.08\\
    MoCo+deletion&67.86&81.43&72.8&81.48&77.84&76.91&69.46&75.40\\
    MoCo+SRL&{68.92}&82.20&{73.67}&81.58&{78.73}&{77.63}&{71.07}&{76.26}\\

    \hline
    \multicolumn{9}{c}{\textit{Continuous Augmentation}} \\
    \hline
    CT-BERT&61.63&76.80&68.47&77.50&76.48&74.31&69.19&72.05\\
    SimCSE-BERT$_{\rm base}$&68.40&82.41&74.38&80.91&78.56&76.85&72.23&76.25\\
    PT-BERT$_{\rm base}$&\bf 71.20&\bf 83.76& \bf 76.34& \bf 82.63& \bf 78.90& \bf 79.42&71.94& \bf 77.74\\
    \toprule[0.8pt]
     \end{tabular}
    \caption{Sentence embedding performance on STS tasks with Spearman's correlation measured. We highlight the highest number for each methods. CLEAR~\cite{wu2020clear} is trained on both English Wikipedia and Book Corpus with 500k steps with their own version of pre-trained models. Result of CT-BERT~\cite{Carlsson2021SemanticRW} is based on the settings of SimCSE~\cite{gao2021simcse}}
    \label{tab:srl-result}
\end{table*}

\begin{table}
\centering
\begin{tabular}{lc}
\hline
\textbf{Models} & \textbf{STS-B dev} \\
\hline
SimCSE-BERT$_{base}$ + None & 82.50 \\
SimCSE-BERT$_{base}$ + Crop & 77.80 \\
SimCSE-BERT$_{base}$ + Deletion & 75.90\\ 
\hline
MoCo-BERT$_{base}$ + None & 82.03 \\ 
MoCo-BERT$_{base}$ + Reorder & 81.89\\
MoCo-BERT$_{base}$ + Duplication & 81.82 \\
MoCo-BERT$_{base}$ + Deletion & 82.97 \\
MoCo-BERT$_{base}$ + SRL & 82.40 \\
\hline 
PT-BERT$_{base}$ & \textbf{84.50}\\
\hline
\end{tabular}
\caption{Results on STS-B development sets. Results of SimCSE~\cite{gao2021simcse} are reported from original paper.}
\label{tab:eval-result}
\end{table}
\paragraph{Model architecture.}
Instead of inputting the same sentence twice to the same encoder, we follow the architecture proposed in Momentum-Contrast (MoCo)~\cite{MoCo} such that PT-BERT can efficiently learn from more negative examples. Samples in PT-BERT are encoded into vectors with two encoders: gradient-update encoder (the upper encoder in Fig.~\ref{fig:architecture}) and momentum-update encoder (the momentum encoder in Fig.~\ref{fig:architecture}). We dynamically maintain a queue to store the sentence representations from momentum-update encoder. 

This mechanism allows us to store as much negative samples as possible without re-computation. Once the queue is full, we replace the "oldest" negative sample with a "fresh" one encoded by the momentum-encoder. 

Similar to the works based on continuous augmentation, at the very beginning of the framework, PT-BERT takes input sentence $s$ and obtains $\bf h_i$ and $\bf h_i'$ with two different encoder functions. We measure the loss function with:
\begin{equation}
     \ell_i = -\log\frac{e^{sim({\bf h_i}, {\bf h_i'})/\tau} }{\sum_{j=1}^M e^{sim({\bf h_i}, {\bf h_{j'}})/\tau}},
     \label{for:loss_our}
\end{equation}
where ${\bf h_i}$ denotes the representations extracted from the gradient-update encoder, ${\bf h_i'}$ represents the sentence embedding in the queue, and $M$ is the queue size. Our gradient-update and momentum-update encoder are based on the pre-trained language model with the same structure and dimensions as BERT-base-uncased~\cite{devlin2018bert}. The momentum encoder will update its parameters similar to MoCo:
\begin{equation}
    \theta_k \leftarrow \lambda\theta_k + (1-\lambda)\theta_q,
    \label{for:moco}
\end{equation}
where $\theta_k$ is the parameter of the momentum-contrast encoder that maintains the dictionary, $\theta_q$ is the query encoder that updates the parameters with gradients, and $\lambda$ is a hyperparameter used to control the updating process.
\paragraph{Relationship with prompt learning.} Rather than directly perform
soft prompting in the embedding space~\cite{li2021prefixtuning, DBLP:journals/corr/abs-2104-06599,liu2021gpt} of the model, our method follows the "plug and play" fashion that project the representations to pseudo sentences only during the period of training. During inference time, PT-BERT predicts the results only with its BERT backbone. Our original intention of designing this procedure is to make the model predict sentence embedding precisely without adding extra computation. In some tasks, fixed-LM tuning~\cite{li2021prefixtuning} in soft prompting becomes competitive only when the language models been scaled to big enough~\cite{lester-etal-2021-power}. While the prompt+LM~\cite{pada,liu2021gpt} tuning adds more burdens for both the period of training and inference. Both prompt+LM and fixed-LM prompt tuning require storing separate copies of soft prompts for different tasks, while our approach only saves the trained BERT model, which draws on some ideas in prompt learning and makes our considerations in computational and memory efficiency and generality.

\section{Experiments}
In this section, we perform the standard semantic textual similarity (STS)~\cite{agirre-etal-2012-semeval,agirre-etal-2013-sem,agirre-etal-2014-semeval,agirre-etal-2015-semeval,agirre-etal-2016-semeval} tasks to test our model. For all tasks, we measure the Spearman's correlation to compare our performance with the previous state-of-the-art SimCSE~\cite{gao2021simcse}. In the following, we will describe the training procedure in detail.

\subsection{Training Data and Settings}
\paragraph{Datasets.} Following SimCSE, We train our model on 1-million sentences randomly sampled from English Wikipedia, and evaluate the model every 125 steps to find the best checkpoints. Note that we do not fine-tune our model on any dataset, which indicates that our method is completely unsupervised.

\paragraph{Hardware and schedule.}
We train our model on the machine with one NVIDIA V100s GPU. Following the settings of SimCSE~\cite{gao2021simcse}, it takes 50 minutes to run an epoch.

\subsection{Implementations}
We implement PT-BERT based on Huggingface transformers~\cite{wolf-etal-2020-transformers} and initialize it with the released BERT$_{base}$~\cite{devlin2018bert}. We initialize a new embedding for pseudo tokens with $128 \times 768$. During training, we create a pseudo sentence $\{0, 1, 2, ..., 127\}$ for every input and map the original sentence to this pseudo sentence by attention. With batches of 64 sentences and an additional dynamically maintained queue of 256 sentences, each sentence has one positive sample and 255 negative samples. Adam~\cite{kingma2014adam} optimizer is used to update the model parameters.  We also take the original dropout strategy of BERT with rate $p = 0.1$. We set the momentum for the momentum-encoder with $\lambda = 0.885$.
\subsection{Evaluation Setup}
We evaluate the fine-tuned BERT encoder on STS-B development sets every 125 steps to select the best checkpoints. We report all the checkpoints based on the evaluation results reported in Table~\ref{tab:eval-result}. The training process is fully unsupervised since no training corpus from STS is used. During the evaluation, we also calculate the trends of alignment-loss and uniformity-loss. Losses were compared with SimCSE~\cite{gao2021simcse} under the same experimental settings. After training and evaluation, we test models on 7 STS tasks: STS 2012-2016~\cite{agirre-etal-2012-semeval,agirre-etal-2013-sem,agirre-etal-2014-semeval,agirre-etal-2015-semeval,agirre-etal-2016-semeval}, STS Benchmark~\cite{cer-etal-2017-semeval} and SICK-Relatedness~\cite{DBLP:conf/lrec/MarelliMBBBZ14}. We report the result of Spearman's correlation for all the experiments. 

\begin{table*}[!ht]
    \centering
    \begin{tabular}{lcccccccc}
    \toprule[0.8pt]
    Method & STS12 & STS13 & STS14 & STS15 & STS16 & STS-B & SICK-R & Avg.\\
    \hline
    \multicolumn{9}{c}{\textit{(a) Ablation studies on pseudo sequence length}}\\
    \hline
    L-64 &67.04&82.04&73.65&81.12&78.64&77.35&71.33&75.88\\
    L-90 &68.94&82.08&74.53&81.22&{79.06}&78.01&71.49&76.48\\
    \bf L-128(Ours) &\bf 71.20&\bf 83.76& \bf 76.34& \bf 82.63& \bf 78.90& \bf 79.42&\bf 71.94& \bf 77.74\\
    L-256 &67.09&82.25&72.63&81.48&78.55&77.30&69.53&75.55\\
    L-360 &68.90&82.21&73.77&81.31&77.50&77.22&69.32&75.75\\
    \hline
    \multicolumn{9}{c}{\textit{(b) Ablation studies on queue size}}\\
    \hline
    Q-192&70.29&\bf 83.78&75.98&82.13&78.48&78.91&72.53&77.44\\
    \bf Q-256(Ours)& 71.20&83.76& \bf 76.34& 82.63& \bf 78.90& \bf 79.42& 71.94& \bf 77.74\\
    Q-320& \bf 71.71&83.36&75.00&\bf 82.99&78.76&79.17&\bf 72.85&77.69\\
    \hline
    \multicolumn{9}{c}{\textit{(c) Evaluations on hard sentence pairs with different length}}\\
    \hline
    SimCSE & 66.54 & 78.50 & 68.76 & 70.27 & 71.31 & - & - & 71.08\\
    PT-BERT & \textbf{72.02} & \textbf{80.24} &\textbf{72.92} & \textbf{74.50} & \textbf{72.50} & - & - & \textbf{74.44}\\
    \toprule[0.8pt]
     \end{tabular}
    \caption{Evaluation results of ablation studies and hard sentence pairs.}
    \label{tab:ablation}
\end{table*}

\subsection{Main Results and Analysis}
We first compare PT-BERT with our baseline: MoCo framework + BERT encoder (MoCo-BERT). MoCo-BERT could be seen as a version of PT-BERT without pseudo token embeddings. Then we apply traditional discrete augmentations such as reorder, duplication, and deletion on this framework. We also compare our work with CLEAR~\cite{wu2020clear} that substitutes and deletes the token spans. Besides, we argue that the performance of these methods is too weak. We additionally propose an advanced discrete augmentation approach that produces positive examples with the guidance of Semantic Role Labeling (SRL)~\cite{srl-orignal, palmer2010semantic} information, instead of random deletion and reordering. SRL-guided augmentation could compensate the errors caused by these factors, acting as a combination of deletion, duplication, and reordering with better accuracy. SRL is broadly used to identify the predicate-argument structures of a sentence, it detects the arguments associated with the predicate or verb of a sentence and could indicate the main semantic information of \textit{who} did \textit{what} to \textit{whom}. For the sentences with multiple predicates, we keep all the sets with order [$\rm ARG0$, $\rm PRED$, $\rm ARGM-NEG$, $\rm ARG1$] and concatenate them into a new sequence. For the sentences without recognized predicate-argument sets, we keep the original sentence as positive examples. In addition to the work based on discrete approaches, we also compare with SimCSE~\cite{gao2021simcse} which continuously augment sentences with ~\textit{dropout}. In Table~\ref{tab:srl-result}, PT-BERT with 128 pseudo tokens further pushed the state-of-the-art results to $77.74\%$ and significantly outperformed SimCSE over six datasets. 

\begin{figure}[t!]
\centering
\subfigure[Alignment loss comparison on STS-B]{
\begin{minipage}[t]{0.5\textwidth}
\centering
\includegraphics[width=1.0\linewidth]{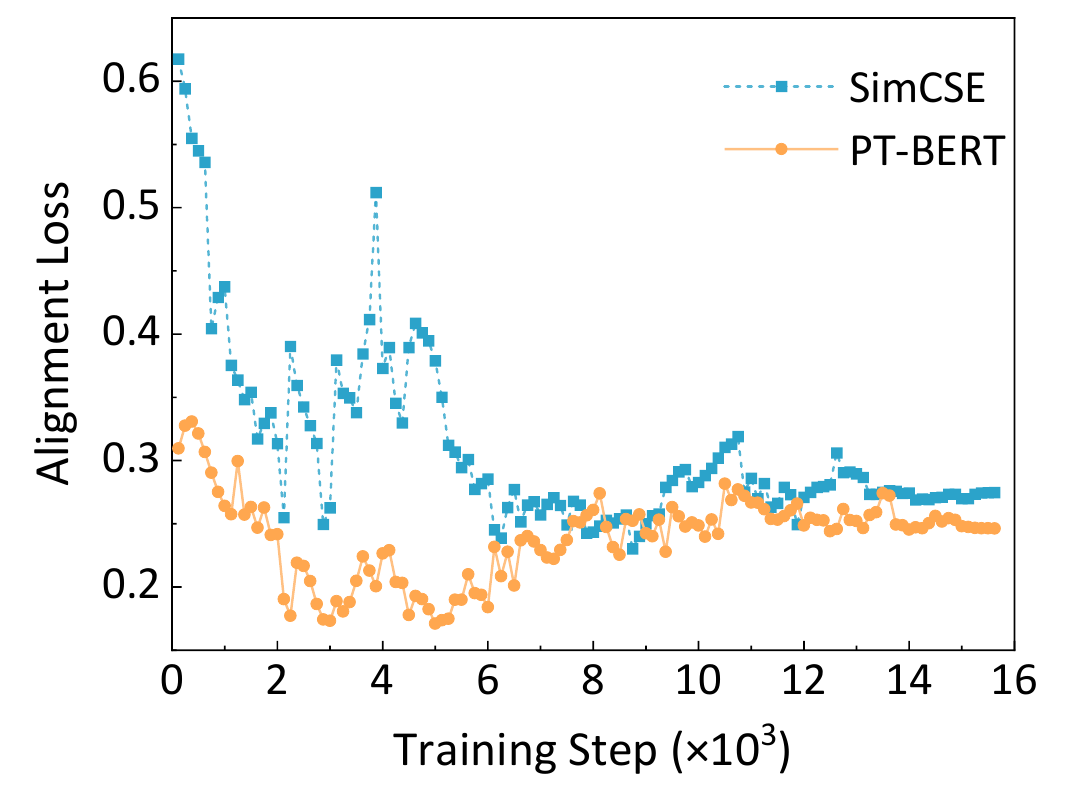}
\end{minipage}
}
\subfigure[Uniformity loss comparison on STS-B]{
\begin{minipage}[t]{0.5\textwidth}
\centering
\includegraphics[width=1.0\linewidth]{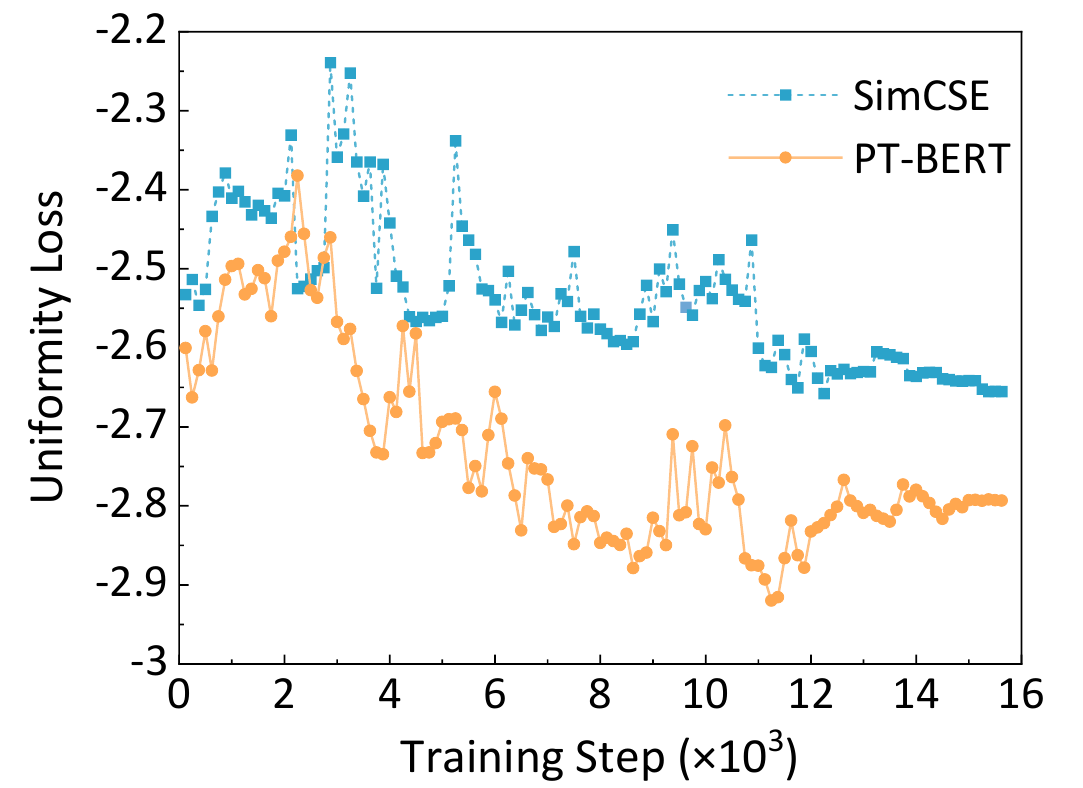}
\end{minipage}
}
    \caption{Alignment and uniformity loss plot for PT-BERT and SimCSE. We visualize the checkpoints every 125 training steps. For both measurements, lower numbers are better.}
    \label{fig:align_unif}
\end{figure}



In Fig~\ref{fig:align_unif}, we observe that PT-BERT also achieves better alignment and uniformity against SimCSE, which indicates that pseudo tokens really help the learning of sentence representations. In detail, alignment and uniformity are proposed by \cite{wang2020understanding} to evaluate the quality of representations in contrastive learning. The calculation of these two metrics are shown in the following formulas:
\begin{equation}
    L_{alignment} = \mathop{E}_{(x, x^+) \sim{p_{pos}}}||f(x) - f(x^+)||^2,
\end{equation}

\begin{equation}
    L_{uniformity} = \log\mathop{E}_{(x, y)\sim{p_{data}}}e^{-2||f(x) - f(y)||^2},
\end{equation}
where $(x, x^+)$ is the positive pair, $(x, y)$ is the pair consisting of any two different sentences in the whole sentence set, $f(x)$ is the normalized representation of $x$.
We employ the final embedding $\textbf{h}$ to calculate these scores.

According to the above formulas, lower alignment loss means a shorter distance between the positive samples, and low uniformity loss implies the diversity of embeddings of all sentences. Both are our expectations for the representations based on contrastive learning. To evaluate our model's performance on alignment and uniformity, we compare it with SimCSE on the STS-benchmark dataset~\cite{cer-etal-2017-semeval}, and the result is shown in Figure~\ref{fig:align_unif}. The result demonstrates that PT-BERT outperforms SimCSE on these two metrics: our model has a lower alignment and uniformity than SimCSE in almost all the training steps, which indicates that the representations produced by our model are more in line with the goal of the contrastive learning.

\section{Analysis}
\subsection{Ablation Studies}
In this section, we first investigate the impact of different sizes of pseudo token embeddings. Then we would like to report the performance difference caused by queue size under the MoCo framework.
\paragraph{Pseudo Sentence Length}
Different lengths of pseudo tokens can affect the ability of the model to express the sentence representations. By mapping the original sentences to various lengths of pseudo tokens, the performance of PT-BERT could be different. In this section, we keep all the parts except the pseudo tokens and their embeddings unchanged. We scale the pseudo sequence length from 64 to 360. Table~\ref{tab:ablation}(a) shows a comparison between different lengths of pseudo sequence in PT-BERT. We find that during training, PT-BERT performs better when attending to pseudo sequences with 128 tokens. Too few pseudo tokens do not fully explain the semantics of the original sentence, 
while too many pseudo tokens increase the number of parameters and over-express the sentence.

\paragraph{Queue Size}
The introduction of more negative samples would make the model's training more reliable. By training with different queue sizes, we report the result of PT-BERT with different performances due to the number of negative samples. In Table~\ref{tab:ablation}(b), queue size $q=4$ performs best. However, the difference in performance between the three sets of experiments is not large, suggesting that the model can learn well as long as it can see enough negative samples.

\subsection{Exploration on Hard Examples with Different Length}
To prove the effectiveness of PT-BERT that could weaken the hints caused by textual similarity, we further test PT-BERT on the sub-dataset introduced in Sec.~\ref{sec:longshortpairs}. We sorted out the positive pairs with a length difference of more than five words and negative pairs of less than two words from STS-(2012-2016). PT-BERT significantly outperforms SimCSE with 3.36$\%$ Spearman's correlation, indicating that PT-BERT could handle these hard examples better than SimCSE. This further proves that PT-BERT could debias the spurious correlation introduced by sentence length and syntax, and focus more on the semantics.

\section{Conclusion} 
In this paper, we propose a semantic-aware contrastive learning framework for sentence embeddings, termed PT-BERT. Our proposed PT-BERT approach is able to weaken textual similarity information, such as sentence length and syntactic structures, by mapping the original sentence to a fixed pseudo sentence embedding. We provide analysis of these factors on methods based on continuous and discrete augmentation, showing that PT-BERT augments sentences more accurately than discrete methods while considering more semantics instead of textual similarity than continuous approaches. Lower uniformity loss and alignment loss prove the effectiveness of PT-BERT and further experiments also show that PT-BERT could handle hard examples better than existing approaches. 

Providing a new perspective to the continuous data augmentation in sentence embeddings, we believe our proposed PT-BERT has great potential to be applied in broader downstream applications, such as text classification, text clustering, and sentiment analysis.

\section*{Acknowledgements}
We would like to thanks the anonymous reviewers for their valuable and constructive comments. This work was supported in part by the Hong Kong RGC grant ECS 21212419, Technological Breakthrough Project of Science, Technology and Innovation Commission of Shenzhen Municipality under Grants JSGG20201102162000001, the Hong Kong Laboratory for AI-Powered Financial Technologies, the CityU Teaching Development Grants under 6000755, and the UGC Special Virtual Teaching and Learning Grants under 6430300.

\bibliography{anthology,custom}
\bibliographystyle{acl_natbib}




\end{document}